\documentclass[letterpaper, 10 pt, conference]{ieeeconf} 
\IEEEoverridecommandlockouts      
\usepackage{times}
\usepackage[pdftex]{graphicx}
\usepackage{graphics}
\usepackage[cmex10]{amsmath}
\usepackage{url}
\usepackage{array}
\usepackage{tabularx}
\usepackage{multirow}
\usepackage{multicol}
\usepackage{booktabs}
\usepackage{epstopdf}
\usepackage{rotating}
\usepackage{csquotes}

\usepackage{amssymb}
\usepackage{amsmath}
\usepackage[all]{xy}    
\usepackage{ifthen}
\usepackage[usenames,dvipsnames]{color}
\usepackage{enumerate}
\usepackage{bm}
\usepackage{siunitx}
\usepackage{xargs}                      
\usepackage[pdftex,dvipsnames]{xcolor}  
\usepackage[colorinlistoftodos,prependcaption,textsize=tiny]{todonotes}
\newcommandx{\unsure}[2][1=]{\todo[linecolor=red,backgroundcolor=red!25,bordercolor=red,#1]{#2}}
\newcommandx{\change}[2][1=]{\todo[linecolor=blue,backgroundcolor=white!25,bordercolor=blue,fancyline,#1]{#2}}
\newcommandx{\info}[2][1=]{\todo[linecolor=OliveGreen,backgroundcolor=OliveGreen!25,bordercolor=OliveGreen,#1]{#2}}
\newcommandx{\improvement}[2][1=]{\todo[linecolor=Plum,backgroundcolor=Plum!25,bordercolor=Plum,#1]{#2}}
\newcommandx{\thiswillnotshow}[2][1=]{\todo[disable,#1]{#2}}

\makeatletter
\let\NAT@parse\undefined
\makeatother
\usepackage[numbers]{natbib}

\newcommand{\secref}[1]{Section~\ref{#1}}
\newcommand{\tabref}[1]{Table~\ref{#1}}

\newcommand{\figref}[1]{Figure~\ref{#1}}

\newcommand*{\Cdot}{\raisebox{-0.25ex}{\scalebox{1.75}{$\cdot$}}}
\newcommand{\myparagraph}[1]{\vspace{0.1in}\noindent\textbf{#1}}

\newboolean{draft-mode}
\setboolean{draft-mode}{true}
\newcommand{\sidenote}[1]{\ifthenelse{\boolean{draft-mode}}{\marginpar{\tiny\raggedright\textsf{\hspace{0pt}#1}}}{}}
\DeclareRobustCommand{\pynote}[1]{\ifthenelse{\boolean{draft-mode}}{\textcolor{green}{\textbf{PY: #1}}}{}}
\DeclareRobustCommand{\arnote}[1]{\ifthenelse{\boolean{draft-mode}}{\textcolor{blue}{\textbf{AR: #1}}}{}}
\DeclareRobustCommand{\nfnote}[1]{\ifthenelse{\boolean{draft-mode}}{\textcolor{red}{\textbf{NF: #1}}}{}}
\DeclareRobustCommand{\mbnote}[1]{\ifthenelse{\boolean{draft-mode}}{\textcolor{cyan}{\textbf{MB: #1}}}{}}

\title{\LARGE \bf More than a Million Ways to Be Pushed.\\ A High-Fidelity Experimental Dataset of Planar Pushing
}
\author{\authorblockN{Kuan-Ting Yu$^1$, Maria Bauza$^2$, Nima Fazeli$^2$, Alberto Rodriguez$^2$} \authorblockA{$^1$
    Computer Science and Artificial Intelligence Laboratory ---
    Massachusetts Institute of Technology\\
    $^2$ Mechanical
    Engineering Department --- Massachusetts Institute of Technology\\
    {\tt\small peterkty@csail.mit.edu}, {\tt\small
      <bauza,nfazeli,albertor>@mit.edu}} \thanks{This work was supported by
    NSF awards [NSF-IIS-1427050] and [NSF-IIS-1551535] through the
    National Robotics Initiative.}
    \includegraphics[width=0.95\textwidth]{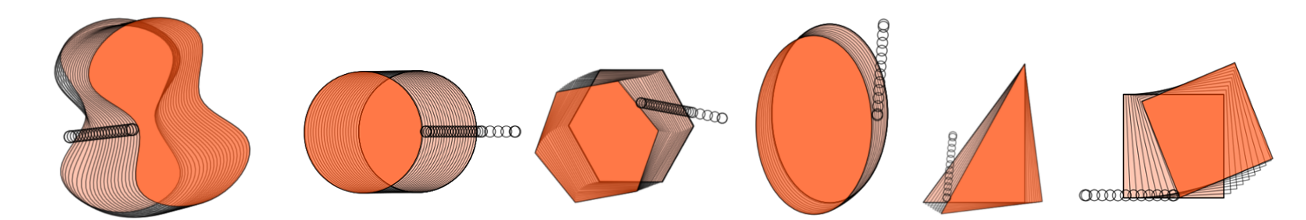} 
}

\begin{document}

\maketitle


\begin{abstract}
  Pushing is a motion primitive useful to handle objects that are too large, too heavy, or too cluttered 
  to be grasped. It is at the core of much of robotic manipulation, in particular when physical interaction is involved.
  It seems reasonable then to wish for robots to understand how pushed objects move.
  
  In reality, however, robots often rely on approximations which yield models that are computable, but also restricted and inaccurate. 
  Just how close are those models? How reasonable are the assumptions they are based on? To help answer these questions, and to get a better experimental understanding of pushing, we present a comprehensive and high-fidelity dataset of planar pushing experiments. 
  The dataset contains timestamped poses of a circular pusher and a pushed object, as well as forces at the interaction.
  We vary the push interaction in 6 dimensions: surface material, shape of the pushed object, contact position, pushing direction, pushing speed, and pushing acceleration.
  An industrial robot automates the data capturing along precisely controlled position-velocity-acceleration trajectories of the pusher, which give dense samples of positions and forces of uniform quality.
  
  We finish the paper by characterizing the variability of friction, and evaluating the most common assumptions and simplifications made by models of frictional pushing in robotics.
\end{abstract}

\section{Introduction}
\label{sec:introduction}

Pushing is a widely used motion primitive for robotic manipulation.
It can aid in the positioning and reorientation of parts \citep{mason1986mechanics, akella1998posing, lynch1996stable, Goldberg1993}; facilitate grasping under pose uncertainty \citep{Brost1988} or clutter \citep{dogar2011framework}; or help in the transportation of large or heavy objects \citep{mericcli2015push, behrens2013robotic}. 
The mechanics of pushing have also been used to aid perception, for example to track the pose of a pushed object \citep{koval2015, jia1999pose, yu2015shape}, to estimate its shape~\citep{yu2015shape}, and to identify inertial parameters such as mass, moment of inertia or coefficient of friction \citep{fazeli2015drop,lynch1993estimating}.
All these applications rely on a good understanding of the mechanics of pushing, which let us predict how an object moves under a certain push.

At an analytical level, pushing is a well understood problem. 
For decades, the mechanics and robotics communities have developed models to explain the interaction at the interface between a pushed object and a support surface~\citep{mason1986mechanics,peshkin1988motion,jia1999pose,liu2011pushing,behrens2013robotic}.
These are usually based on Coulomb's friction law, often rewritten
as the maximum-power inequality~\citep{Goyal1991}. In \secref{sec:relatedWork}, we summarize these, and see how they have led to compact and deterministic models that, under sufficient assumptions, can be used to explain and control the motion of a pushed object. 

The reality, however, is bitter. Predicting the motion of a pushed object is not trivial. In practice, the sensitivity of the task to small changes in contact geometry, along with the variability of friction, hinders accurate predictions.

More recently, data-driven models \citep{salganicoff1993vision, mericcli2015push, lau2011automatic, walker2008pushing, zhou2016convex} have been proposed as an alternative approach to analysis. Studies are still incipient and either do not offer sufficient generality, or do not address variability.
%
A lack of common datasets or benchmarks may explain why learning has not yet had the same effect that it has in other disciplines, such as computer vision. Datasets could facilitate research in model development by enabling evaluation and comparison of solutions.  
Although robots excel at accuracy and repetition, capturing large amounts of real data requires  setting and resetting of experiment conditions, which is tedious with human intervention, and difficult without it. This is in stark contrast, for example, to collecting digitized daily images for computer vision research \cite{torralba200880}.


In this study, we have automated the setting and execution of controlled pushing experiments, and captured a large high-fidelity dataset of pushing interactions. The dataset, detailed in \secref{sec:dataSet}, includes a wide variety of controlled pushes recorded at a high sample rate with both a force/torque sensor, and a Vicon tracking system. 





The dataset covers variations in surface material, object shape,  contact position, pushing direction, pushing speed, and pushing acceleration. To our knowledge, this is the first dataset of this caliber. A significant novelty of the dataset is that it contains dynamic pushing where the inertial components have an appreciable effect over frictional forces, for which there is little work in the literature. We hope it will provide a tool for an experimental study of pushing.

In summary, our contributions are:
\begin{itemize}
    \item A dataset of planar pushing containing time series of high-fidelity poses of both the pusher and the pushed object, and forces experienced by the pusher. This yields more than a million timestamped data points of poses and forces for each combination of shape and material.
    \item An evaluation of common assumptions made by analytical models of pushing in robotics. 
\end{itemize}


\begin{figure}
  \begin{center}
    \includegraphics{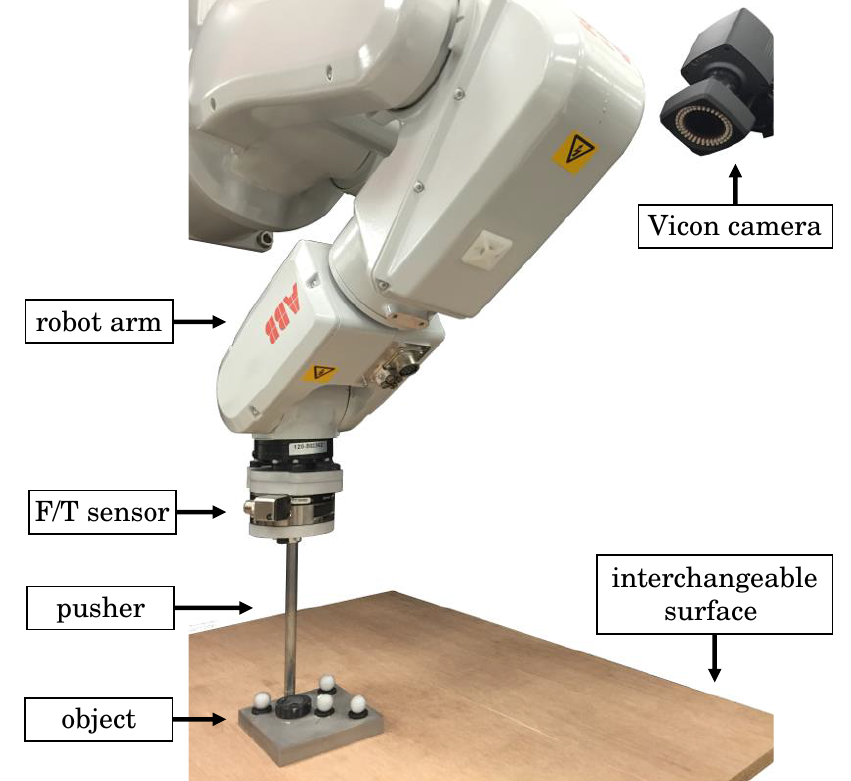}
  \end{center}
  \caption{Data capturing hardware: A steel pusher is attached to an ATI force/torque sensor, and driven by an ABB IRB 120 robot arm. The object (stainless steel block) is instrumented with reflective 
  markers and tracked with a Vicon motion tracking system. The block slides on top of an interchangeable support surface.}
  \label{fig:hardware_robot}
\end{figure}

\section{Related Work}
\label{sec:relatedWork}

One of the most common assumptions in robotic pushing, and possibly in robotic manipulation, 
is quasistatic interaction. In the context of pushing, quasistatic interaction means that the velocity of the involved objects is small enough that inertia is negligible. Instantaneous motion is then a consequence of the balance between contact forces, frictional forces, and gravity. The quasistatic assumption makes the problem more tractable, yielding simpler models, and is a reasonable assumption for the scales and speeds in much of robotic manipulation~\cite{mason2001mechanics}. 

\citet{mason1986mechanics} starts the line of work on pushing by proposing a voting theorem to determine the rotation direction of a pushed object. \citet{Goyal1991} propose a limit surface representation to map motions and frictional forces of a sliding object. These serve as the foundation of much subsequent work on pushing. \citet{lee1991fixture} propose to approximate the limit surface as an ellipsoid to improve computational time.
\citet{lynch1992manipulation} apply the ellipsoidal approximation to derive a closed-form analytical solution for quasistatic pushing, including both sticking and sliding behaviors. \citet{howe1996practical} explore more approximation methods of limit surfaces, and provide a guide for choosing between them based on the pressure distribution, computation cost and accuracy.  
These models provide functional representations of Coulomb's friction law, and the maximum power inequality~\citep{Goyal1991}. We find practical use cases, for example,
for stable pushing of a planar object with a fence-shaped finger \citep{lynch1996stable}, 
for planning robust push-grasps for objects in clutter \citep{dogar2011framework},
and for planning in-hand manipulation with patch contacts \citep{ChavanDafle2015a}.

\citet{peshkin1988motion} address the uncertainty in pushing by sampling all possible pressure distributions and predicting a range of possible object motions. \citet{jia1999pose} investigate dynamic pushing but assume frictionless interaction between pusher and object. \citet{behrens2013robotic} instead studies dynamic pushing assuming infinite friction between pusher and object. \tabref{tab:assump} summarizes these works along with their assumptions and approximations. In this paper we explicitly validate the assumptions marked with an asterisk (*).

\begin{table}
  \caption{Assumptions and approximations made in prior work}
  \label{tab:assump}
  
  \centering
  \begin{tabular}{|l|l|} \hline
    \bfseries Condition       & \bfseries Work examples    \\ \hline
    Uniform friction*      & [all] \\ \hline
    Known pressure distribution      & \citep{Goyal1991, lynch1992manipulation, lee1991fixture, dogar2011framework} \\ \hline
    Known center of friction  & \citep{mason1986mechanics} \\ 
    (the centroid of pressure distribution)  &   \\ \hline
    Coulomb friction*          & \citep{mason1986mechanics, lynch1992manipulation, jia1999pose, lynch1996stable, ChavanDafle2015a, dogar2011framework} \\ \hline
    Maximum power inequality*  & [all] \\
    (generalized Coulomb friction)  &  \\ \hline
    Quasistatic interaction              & \citep{lynch1992manipulation, mason1986mechanics, lynch1996stable, dogar2011framework} \\ \hline
    Ellipsoidal limit surface* & \citep{lee1991fixture, lynch1992manipulation}  \\ \hline
    Sliding (frictionless) pushing      & \citep{jia1999pose} \\ \hline
    Sticking (infinite friction) pushing          & \citep{behrens2013robotic} \\ \hline
    \multicolumn{2}{l}{* Conditions that we explicitly validate in this paper.}
  \end{tabular}
  
\end{table}

\section{The Pushing Dataset}
\label{sec:dataSet}


\begin{table*}
  \caption{Summary of dimensions explored in the dataset.}
  \label{tab:Dimensions}
	\centering
	\begin{tabular}{|l|l|}
          \hline
          \bfseries Shape & 
            \includegraphics[width=0.22\linewidth]{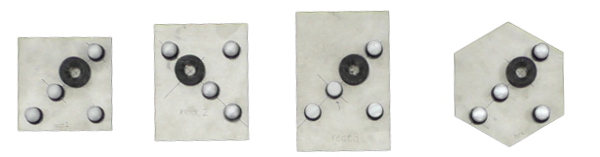}
            \includegraphics[width=0.25\linewidth]{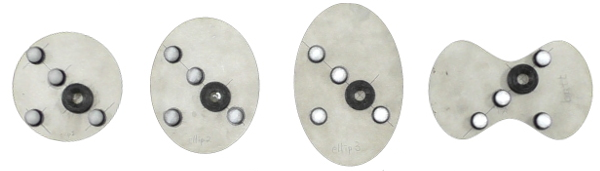} 
            \includegraphics[width=0.13\linewidth]{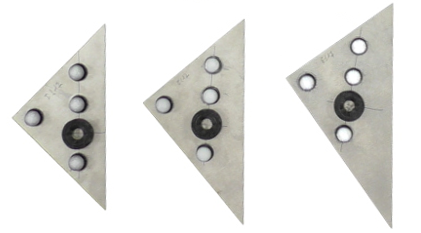} 
          
          \\
           & \texttt{rect1}, \texttt{rect2}, \texttt{rect3}, \texttt{hex}, \texttt{ellip1}, \texttt{ellip2}, \texttt{ellip3} ,   \texttt{butter} ,  \texttt{tri1}, \texttt{tri2}, \texttt{tri3}\\ \hline
          \bfseries Surface  & \texttt{abs}, \texttt{derlin}, \texttt{polywood}, \texttt{pu} \\ \hline
          \bfseries Speed (mm/s) & 10, 20, 50, 75, 100, 150, 200, 300, 400, 500 \\ \hline
          \bfseries Acceleration (ms$^{-2}$) & 0, 0.1, 0.2, 0.5, 0.75, 1, 1.5, 2, 2.5\\ \hline
          \bfseries Initial contact & 33 points for \texttt{tri1-3} and \texttt{hex}, 40 for \texttt{ellip1-3} and \texttt{butter}, and 44 for \texttt{rect1-3} \\ \hline
          \bfseries Initial push direction & 0$^\circ$, 20$^\circ$, 40$^\circ$, 60$^\circ$, 80$^\circ$, -20$^\circ$, -40$^\circ$, -60$^\circ$, -80$^\circ$   \\ \hline
    \end{tabular}
\end{table*}

This dataset records the poses of a pusher and a pushed object together with the interaction forces for a variety of pushing experiments. Variations in experiments cover 6 dimensions: object shape, surface material, pusher direction, pusher speed, pusher acceleration, and initial contact position. This section describes these dimensions, which are also summarized in \tabref{tab:Dimensions}. For extended details, refer to the dataset website \cite{labwebsite}:

\begin{itemize}

\item[$\Cdot$] \textbf{Shape.} Different shapes can give us insights into phenomena such as the dependence of friction with variations in the support pressure distribution. We use 3 rectangles with different aspect ratios (\texttt{rect1-3}), 3 right triangles with different skews (\texttt{tri1-3}), 3 ellipses with different eccentricities (\texttt{ellip1-3}), 1 hexagon (\texttt{hex}), and 1 butterfly shape (\texttt{butter}). See \tabref{tab:objects} for dimensions and other physical properties. All objects are fabricated in stainless steel, and bead blasted to give a rough finish free of burrs.
\item[$\Cdot$] \textbf{Surface material.} The support surface where the object slides is of great importance as it dictates the frictional interaction with the object. We experiment with four surfaces: i) ABS, ii) Delrin, iii) plywood, and iv) polyurethane (hardness 80A durometer). The first two are widely used hard plastics. The third is a softer material and the fourth has a rubber-like texture. We will refer to the materials as \texttt{abs}, \texttt{delrin}, \texttt{plywood}, and \texttt{pu} respectively throughout the paper. In section \ref{sec:CharacFriction} and \ref{sec:EvalClassicalConcept}, we characterize relevant frictional properties of these surfaces. 
%

%
\item[$\Cdot$] \textbf{Speed.} Speed dictates the regime in which the object moves: quasistatically (negligible inertia) or dynamically (meaningful inertia). We explore pusher trajectories with constant speeds: 10, 20, 50, 75, 100, 150, 200, 300, 400, and 500 mm/sec.
\item[$\Cdot$] \textbf{Acceleration.} The acceleration of the pusher is a relatively unexplored dimension. We capture interactions both with constant speed (zero acceleration) and with constant accelerations 0.1, 0.2, 0.5, 0.75, 1, 1.5, 2, and 2.5 ms$^{-2}$ starting from rest.
\item[$\Cdot$] \textbf{Contact position.} Each object is pushed at a number (between 33 and 44) of evenly-spaced contact locations.
\item[$\Cdot$] \textbf{Push direction.} For each contact location we vary the direction of the push between -80$^\circ$ to 80$^\circ$ around the contact normal with increments of 20$^\circ$, for a total of 9 directions.
\end{itemize}

Each experiment executes an open-loop pusher trajectory in the reference frame of the initial pose of the object, leading to evolving contact geometry between object and pusher. The trajectory is executed and recorded at 250~Hz, which allows us to explore different contact interactions efficiently, including transitions between sticking and sliding.  

The experiments are position controlled in preference to force controlled for three reasons: First, the position, speed and acceleration of the pusher can be controlled very accurately with an industrial robot, whereas force sensors typically have much lower signal-to-noise ratio; Second, controlling the force between pusher and object is challenging because it is constrained by friction, and errors in the friction coefficient can lead to unexpected trajectories; Third, although we do not control force directly, we record the force and pose of the object at a high frame rate. The data is still useful to study the relation between forces and motions.

\section{Data Collection Specifications}
\label{sec:collection}

In this section we detail the data collection system and the automated process designed to record the pushes. \figref{fig:hardware_robot} shows the setup: a 6 DOF industrial robotic manipulator equipped with a stiff cylindrical rod acting as a pusher.
	
\subsection{Hardware}

\myparagraph{Robot.}
The system uses an ABB IRB 120 industrial robotic arm with 6 DOF to control precisely the position, velocity
and acceleration of its tool center point (TCP). The robot has a horizontal reach of 580 mm and a payload of 3 kg, which is sufficient since the pushed objects have a mass in the order of 1 kg. 

\myparagraph{Force sensing.}
We use an ATI Gamma force-torque sensor rigidly attached to the 6th link of the robot to measure the reaction force from the object on the pusher. The sensor has high sensitivity with force resolution of 1/160 N in the pushing plane, and torque resolution of 1/2000 N$\cdot$m perpendicular to the pushing plane. 

\myparagraph{Motion sensing.}
We track the pose of the object with a Vicon motion tracking system, composed of 5 Bonita cameras with a wide field of view. Each object is fitted with 4 reflective markers. Although 3 are in theory sufficient, in practice 4 asymmetric markers give more stable readings. 
	

The noise in the recording system is quite small. The accuracy of the object position depends on the accuracy of the Vicon system, which is below 0.5 mm for translation and 0.5$^\circ$  for rotation. The pose of the pusher is directly given by the robot, with an accuracy of 0.1 mm. 

\myparagraph{Pusher.}
The robot is equipped with a stiff cylindrical steel pusher, mounted on and perpendicular to the measurement plate of the force-torque sensor. The pusher has length 156~mm and diameter 9.5~mm, which we found to be a good trade-off to minimize occlusions and provide rigidity. 

\myparagraph{Objects.}
We use a total of 11 objects, all water-jet cut in stainless steel for durability. They are bead blasted to remove burrs, retaining a more realistic ``rough'' surface. Object mass ranges between 0.75 and 1.4 kg depending on the shape. All objects are 13 mm thick. The friction coefficient between the pusher and the object is approximately 0.25, which was determined using a traditional variable slope experiment.


\begin{table}
  \caption{Set of objects in the dataset. Physical properties.}
  \label{tab:objects}
	\centering
	\begin{tabular}{|l|l|l|l|}
          \hline
          \bfseries Object & \bfseries Mass (g) & \bfseries Dimension (mm) & \bfseries Moment of\\
           & &  & \bfseries inertia (g$\cdot$m$^2$)\\
          \hline
          \texttt{rect1}  & 837    & w:90, h:90               & 1.13 \\ \hline  
          \texttt{rect2}  & 1045   & w:90, h:112.5            & 1.81 \\ \hline
          \texttt{rect3}  & 1251   & w:90, h:135              & 2.74 \\ \hline
          \texttt{hex}    & 983    & circumradius: 60.5       & 1.50\\ \hline
          \texttt{ellip1} & 894    & w:105, h:105             & 1.23\\ \hline
          \texttt{ellip2} & 1110   & w:105, h:130.9           & 1.95\\ \hline
          \texttt{ellip3} & 1334   & w:105, h:157             & 2.97\\ \hline
          \texttt{butter} & 1197   & w1:95.3, w2:54.7, h: 156 & 2.95\\ \hline  
          \texttt{tri1}   & 803    & leg1: 125.9, leg2: 125.9 & 1.41\\ \hline
          \texttt{tri2}   & 983    & leg1: 125.9, leg2: 151.0 & 2.11\\ \hline
          \texttt{tri3}   & 1133   & leg1: 125.6, leg2: 176.5 & 2.96\\ \hline
\end{tabular}
\end{table}

\subsection{Software}
	
To facilitate integration of various components such as robot control, force-torque sensor, and motion tracker, we use the Robot Operating System (ROS) framework. Data streams (robot pose, object pose and force-torque) are published as ROS topics and recorded at 250~Hz. The experiments are logged as ROS bag files and parsed into HDF5 and JSON format. Refer to \cite{labwebsite} for more format details.




\subsection{Data Collection Process}

The pushing experiments follow these steps:
\begin{itemize}
\item[1.] The tracker locates the object.
\item[2.] The robot executes an open-loop straight push along a predefined position-velocity-acceleration trajectory, in the initial reference frame of the object. The Vicon tracker and force-torque sensor record the interaction.
\item[3.] If needed, the robot resets the location of the object by dragging it to approximately the center of the plate.
\item[4.] Iterate.
\end{itemize}
The reset mechanism, key for capturing a very large number of experiments, is implemented by a thick tapered washer on the top of the object, that allows the pusher to easily drag the object in the plane.

The pusher starts in contact with the object and follows a straight line of 5 cm. The figure under the paper title shows 6 examples of straight-line pushes.

The data collection results in an approximate total of 6,000 pushes per object and surface. Each push produces an average of 200 timestamped interactions. In total, the experiments yield more than a million triples of pusher motion, object motion and pushing force.



\section{Variability of Dynamic Surface Friction}
\label{sec:CharacFriction}
To support the previous dataset, we have conducted a series of experiments to characterize surface friction. In particular we are interested in studying the variability of the effective coefficient of friction with respect to these factors of a sliding motion:

\begin{enumerate}
\item location,
\item repetition,
\item speed, and 
\item direction.
\end{enumerate}

\begin{figure}[t]
  \begin{center}
    \includegraphics[width=\linewidth]{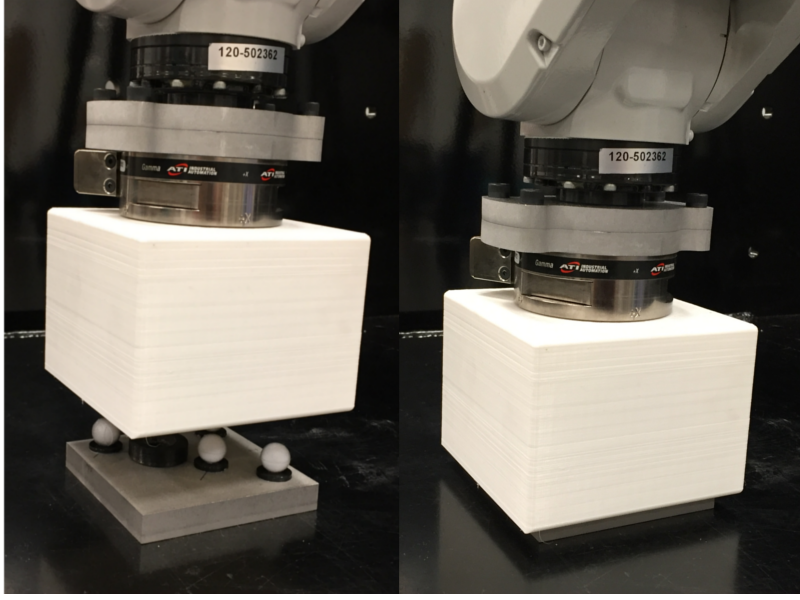}
  \end{center}
  \caption{Cage for experiments of variability of surface friction. When engaged (right) the object fits loosely in the cage. The force-torque sensor measures both the force and moment of friction in the plane.}
  \label{fig:cage_finger}
\end{figure}

To study these effects we design the cage in \figref{fig:cage_finger} to push the rectangle \texttt{rect1} in a controlled manner. The cage does not clamp the object but traps it, with a small gap ($<$ 1 mm) between object and cage. This gives the robot full control over the object sliding motion. We are interested in characterizing the dynamic friction force between the object and the surface, measured as the reaction force in the horizontal plane on the force/torque sensor at the robot wrist. We define then the theoretical dynamic coefficient of friction (DCoF) as the ratio between the measured reaction force $f_f$ and the supporting normal force $f_n$, $\textnormal{DCoF}=\frac{f_f}{f_n}$.

In the experiments, the robot performs line scanning motions tailored to exploring specific dimensions: 1) location, 2) repetition, 3) speed and 4) direction. For dimensions 1-3) in each pass of the scan, the robot pushes the block from left to right and back to the starting point. It then moves down to transit to the next scanline. Neighboring scanlines are separated by 5 mm for 1) for high spatial resolution, and 100 mm for 2) and 3) for non-overlapping scans. All scanlines for 4) follow the diameters of a circle at the surface center.  
Data for 1), 2), and 4) are captured at a speed of 20 mm/s; for 3), we conduct 10 scans for each of the 10 speeds described in \secref{sec:dataSet}.

\begin{figure*}
  \begin{center}
    \includegraphics{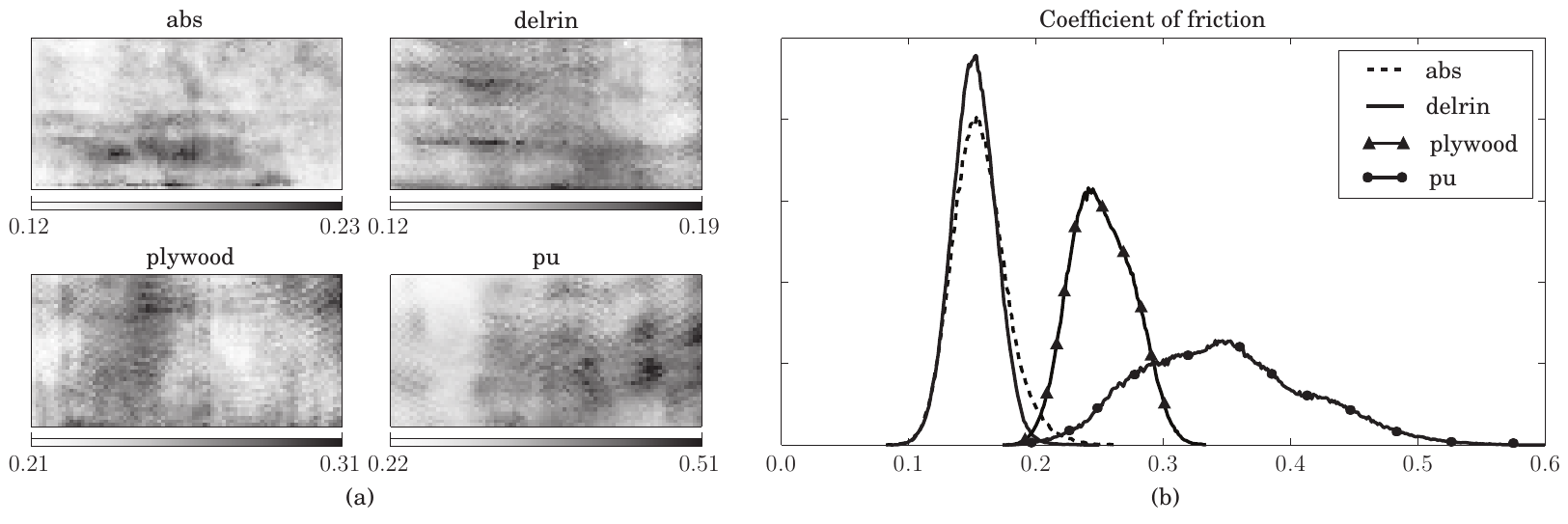}
  \end{center}
  \caption{a) Spatial distribution of the coefficient of friction (DCoF) for four materials. The darker the color, the higher the coefficient. b) Histogram of the same distributions.}
  \label{fig:friction_over_location}
\end{figure*}

\myparagraph{1) Spatial variability.}
Surface imperfections yield variation in the DCoF. \figref{fig:friction_over_location}a shows the spatial distributions recovered for all four surfaces. The areas mapped are approximately 20 cm by 40 cm. It is interesting to note that even seemingly smooth and uniform materials such as \texttt{delrin} have distinguishable differences on the surface. \figref{fig:friction_over_location}b shows the histogram of the measured DCoF for each surface material. Sorting their standard deviation from low to high, we have \texttt{delrin}: 0.016, \texttt{abs}: 0.017, \texttt{plywood}: 0.024, and much larger \texttt{pu}: 0.064. Interestingly, the histograms resemble Gaussian distributions which could be considered as a basic model for frictional sliding.



\begin{figure}
  \begin{center}
    \includegraphics[width=3.4in]{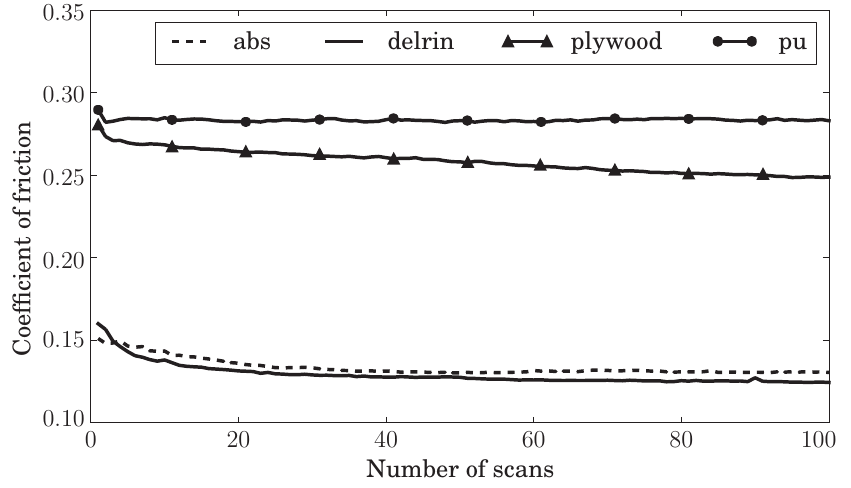}
  \end{center}
  \caption{Evolution of the coefficient of friction (DCoF) over 100 scans, for four different materials. Note that \texttt{abs} and \texttt{delrin} have a relatively short break-in phase, \texttt{plywood} does not stop degrading, and \texttt{pu} is more resistant to abrasion.}
  \label{fig:friction_change_over_time_lineplot}
\end{figure}

\myparagraph{2) Temporal variability.}
A surface generally becomes smoother after being repeatedly rubbed in a polishing process. Similarly, sliding objects polish the surface they slide on and therefore change its effective DCoF. Here we quantify the polishing effect on newly purchased surfaces. 
\figref{fig:friction_change_over_time_lineplot} shows a decreasing trend of the effective DCoF for all materials. This effect is sometimes called break-in. After 100 scans, their respective DCoFs change like:

\begin{itemize}
\item \texttt{abs}: 0.15 to 0.13 (-13.6\%);
\item \texttt{delrin}: 0.16 to 0.12 (-22.2\%);
\item \texttt{plywood}: 0.28 to 0.24 (-11.3\%);
\item \texttt{pu}: 0.29 to 0.28 (-2.3\%).  
\end{itemize}

We observe that \texttt{delrin} and \texttt{abs} have an appreciable break-in period after which the DCoF converges to an almost constant value. For \texttt{plywood}, the break-in period is much longer. For \texttt{pu}, the break-in period is almost non-existent, hinting that for the range of forces we consider, there is almost no degradation of the material over time. Indeed, the type of polyurethane we used is abrasion-resistant. 

\begin{figure}
  \begin{center}
    \includegraphics[width=3.4in]{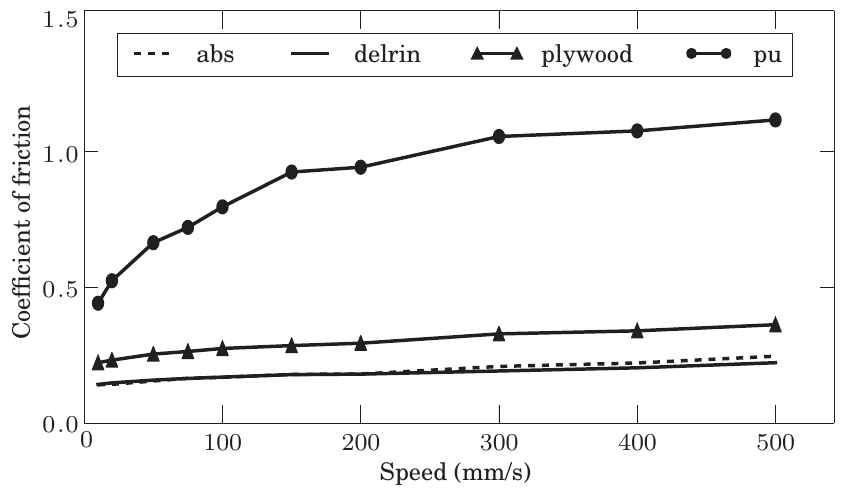}
  \end{center}
  \caption{Change of the coefficient of friction (DCoF) with sliding speed of the object.}
  \label{fig:friction_change_over_speed}
\end{figure}

\myparagraph{3) Speed variability.} 
Coulomb friction states that the magnitude of the friction force should not depend on the object sliding speed. \figref{fig:friction_change_over_speed} shows the results for experiments conducted with different speeds. Indeed, \texttt{delrin}, \texttt{abs} and \texttt{plywood} present little variability of DCoF with speed. The DCoF of \texttt{pu} however, increases up to 1.0 for high speeds. The phenomenon is already observed in \cite{roth1943frictional} for rubbers, and \cite{clemitson2015castable} states that \texttt{pu} possess this characteristic. Coulomb friction then would not be a good approximation when the speed of experiments spans a wide range. 

\begin{figure}
  \begin{center}
    \includegraphics[width=3.4in]{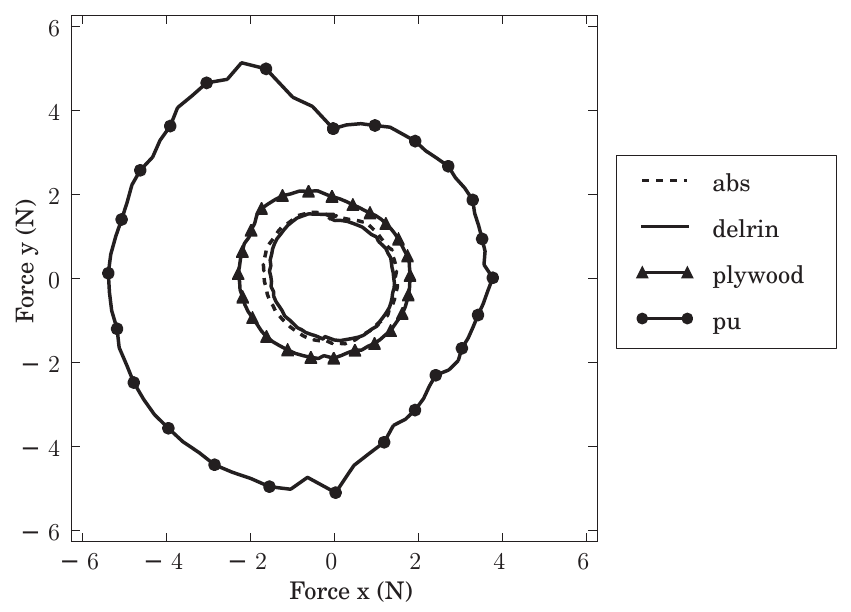}
  \end{center}
  \caption{Directionality of friction force. Experiments show that \texttt{abs}, \texttt{delrin}, and \texttt{plywood} are mostly isotropic, i.e., friction is not direction dependent.}
  \label{fig:friction_change_over_direction}
\end{figure}

\myparagraph{4) Direction variability.} 
When a material presents friction independent of the sliding direction, we say it is isotropic; otherwise, anisotropic. To test it, we perform successive scans where we force the object to slide through the center of the plate in different directions. \figref{fig:friction_change_over_direction} shows the set of friction forces collected. An isotropic material would show a circular force profile. The figure shows that \texttt{abs} and \texttt{delrin} are close to isotropic, \texttt{plywood} slightly less, and \texttt{pu} the least. For \texttt{pu}, the ratio between the largest friction and the smallest is around 3/2, which is a significant difference. This could explain, in part, the large standard deviation of the DCoF observed in \figref{fig:friction_over_location}b since scans are run forward and backward.

\section{Evaluation of Models of Frictional Sliding}
\label{sec:EvalClassicalConcept}

In this section we study whether:  
\begin{enumerate}
\item frictional sliding follows the maximum-power inequality, a.k.a. maximum dissipation principle;
\item the limit surface of a particular material can be well approximated by an ellipsoid.
\end{enumerate}

We conduct experiments with the same setup as in \secref{sec:CharacFriction}, with all scans passing through the center of the plate. To analyze the behavior of the frictional sliding wrench (force and torque), we conduct experiments controlling the instantaneous sliding twist of the object (linear and angular velocity) as it passes through the center of the plate. 

To achieve this we generate trajectories with different ratios of translation and rotation velocity. We perform linear scans that approach the center of the plate at different angles in increments of $5^\circ$ and from a starting distance from the center $\in\left\{50,25,12.5,0\right\}$ mm. For each scan, the object rotates between angles $\theta$ and $-\theta$ where we vary $\theta\in\left\{-88^\circ\ldots 88^\circ\right\}$ with increments of $4^\circ$.

\myparagraph{1) Principle of maximum-power inequality.}
The curves in \figref{fig:friction_change_over_direction} are known as limit curves (LC), i.e., the set of all possible frictional forces between object and surface material in pure translational sliding. 
The principle of maximum-power inequality \cite{Goyal1991} states that the resolution of frictional force and sliding motion is such that dissipation of power will be maximized. We can state the principle as:
\[\forall f^* \in \textrm{LC}, (f-f^*) \cdot v \geq 0,\]
where $f$ and $v$ are the friction force and sliding velocity at contact, and $f^*$ is any other friction force in the LC. 

In a general contact/friction problem, this principle is difficult to resolve, since it is a constraint that involves both forces and motions~\citep{ChavanDafle2015a}. In our experiments, however, we force a particular velocity on the object. Then it is straightforward to verify if
%
$\Delta P=f \cdot v - \max_j (f_j \cdot v) \geq 0$, where $j$ spans all points in the LC.
%
To avoid issues with the different types of frictional variability discussed \secref{sec:CharacFriction}, we only use data for the object passing through a particular point of interest.

\begin{figure}
  \begin{center}
    \includegraphics[width=3.4in]{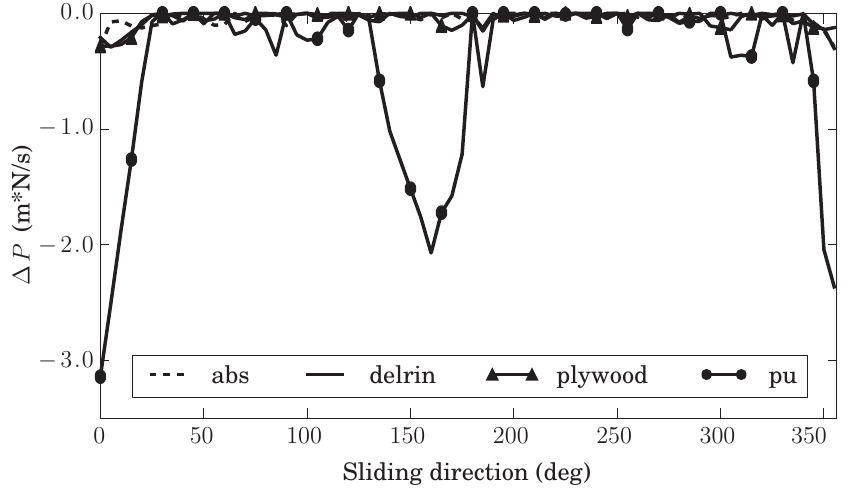}
  \end{center}
  \caption{Difference between power dissipated and maximum dissipable power ($\Delta P$) for the squared object \texttt{rect1} sliding along different directions. The maximum power inequality expresses that $\Delta P$ should be zero. }
  \label{fig:friction_maximal_dissipation_overmaterial}
\end{figure}

\figref{fig:friction_maximal_dissipation_overmaterial} shows $\Delta P$ for experiments with different direction when passing through the center of the plate. All materials except for \texttt{pu}, yield $\Delta P$ very close to 0. For \texttt{pu}, there are 2 regions where $\Delta P$ is significantly less than 0. They correspond to the abrupt transitions at the top and bottom of its limit curve in \figref{fig:friction_change_over_direction}.

\begin{figure}
  \begin{center}
    \includegraphics[width=2.75in]{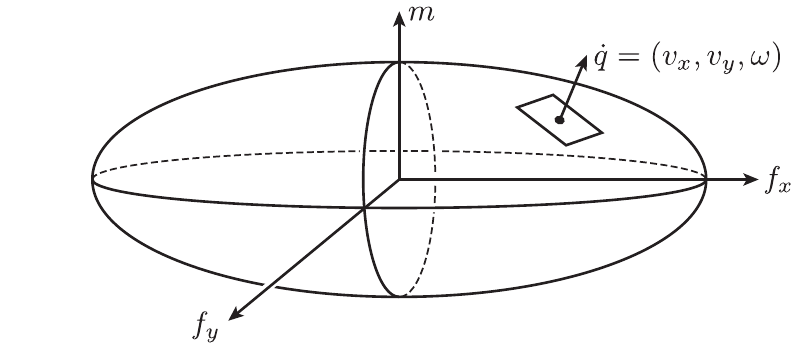} \\
    {\footnotesize (a)} \\
    \includegraphics{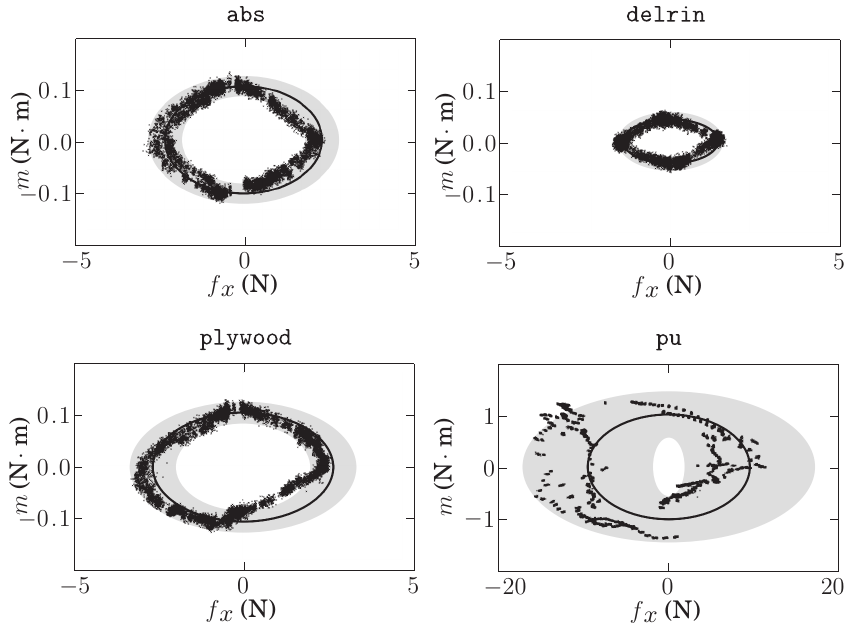} \\
    {\footnotesize (b)}
  \end{center}
  \caption{(a) Conceptual limit surface. The set of all possible frictional wrenches (force/torques) that the sliding object can receive from the surface (b) Experimental set of frictional force/torques exerted on the sliding object \texttt{rect1} by different materials, and best ellipsoidal fitting, in the  $f_x$-$m$ plane.}
  \label{fig:friction_limitsurface_ellipsoid}
\end{figure}

\myparagraph{2) Limit surface and ellipsoidal approximation.}
A sliding planar object can both translate and rotate. The corresponding frictional wrench will have then both force and torque components. The limit curve (LC) discussed above  generalizes into a limit surface (LS), the 2D set of frictional wrenches in the 3D wrench space that a surface can exert on a sliding object. \figref{fig:friction_limitsurface_ellipsoid}a shows a simple visualization of that limit surface. 
Similar to the LC, the LS 
works as follows: If the object is sliding, the friction wrench lies on the LS; otherwise, it lies strictly inside the LS. The maximum-power inequality dictates then that the motion corresponding to a particular frictional force must be orthogonal to the LS at that point~\citep{Goyal1991}. 

For computational reasons, the LS is occasionally approximated as an ellipsoid~\citep{howe1996practical}. Here we verify that approximation by constructing the real limit surface from real measurements. \figref{fig:friction_limitsurface_ellipsoid}b shows the recovered LS for four materials. We fit an ellipsoid to that data, by assuming it is centered at the origin, and estimating the moment magnitude from pure rotational motion and force magnitude from pure translational motion. The shade region shows the $2\sigma$ uncertainty region. 

Observe that the real limit surface is closer to thicker noisier ring. We can also see that the underlying curve of the data  resembles an ellipse but not exactly. Finally, we observe that \texttt{delrin} has the most symmetric LS, \texttt{abs} and \texttt{plywood} are slightly biased toward the left side, possibly due to slight anisotropy, and \texttt{pu} resembles very little to an actual LS.

\section{Stochasticity of pushing motion}
Experiments in \secref{sec:CharacFriction} show that friction expresses a degree of variability in several dimensions, including space, time, speed and direction. If we want to use simple models, that will not be overconfident about frictional behavior, one could treat friction as a stochastic process. Here, we describe how the uncertainty looks like and motivate further in-depth studies to gain insight about the following questions:
\begin{itemize}
    \item Does the object motion distribution appear Gaussian?
    \item How wide is it?
    \item Are the predictions from models in use close to the experimental behavior?
    \item Is the distribution dependent on the surface material?
\end{itemize}

\begin{figure*}
  \begin{center}
    \includegraphics{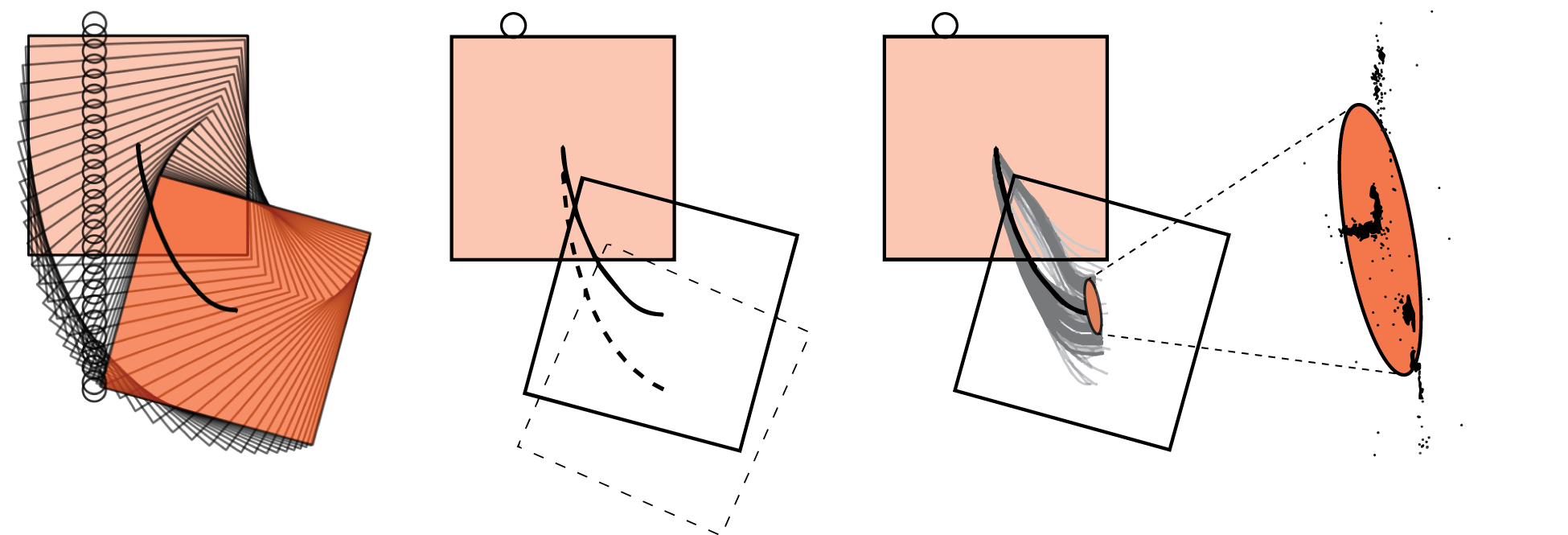}\\
    {\footnotesize(a) \qquad\qquad\qquad\qquad\qquad\qquad\qquad (b)  \qquad\qquad\qquad\qquad\qquad\qquad\qquad\qquad\qquad\qquad (c) \qquad\qquad\qquad\qquad}
 \end{center}
  \caption{Example of 2000 pushes of object \texttt{rect1} on surface material \texttt{abs}. (a) Mean object trajectory. The thick solid line traces the center of mass of the object. (b) Comparison with simulated trajectory (dashed line) with the model in \cite{lynch1992manipulation}. The experiment yields a significant difference. (c) Distribution of the final locations of the center of mass of the object. Note the multi-modality.}
  \label{fig:rep_push_viz_abs}
\end{figure*}

To do so, we repeat a particular straight-line push experiment 2,000 times. The particular settings are:
\begin{itemize}
    \item shape: \texttt{rect1};
    \item contact location: half way in between the center and edge of the block side;
    \item contact angle: normal direction;
    \item speed: 20 mm/s (quasistatic speed);
    \item acceleration: 0 mm/s$^{2}$;
    \item surface: all 4 materials;
    \item pusher displacement: 15 cm.
\end{itemize}
For parsing the results, we denote the object trajectory as starting from $(x,y,\theta)=(0,0,0)$, and ending at $(\Delta x, \Delta y, \Delta \theta)$.  \figref{fig:rep_push_viz_abs} shows the resulting trajectories. The distribution of final poses seems to have at least three modes, and its shape is clearly not Gaussian. \tabref{tab:rep_push} shows the standard deviation (std) of ending poses, which depends on the surface type. We normalize the error by the mean displacement to get an error rate (\%). Qualitatively the standard deviation is related to the characterization of friction variability in \secref{sec:CharacFriction}. It would be interesting to further investigate what is the nature of that relationship. 

\begin{table}
  \caption{Distribution of object displacements after repeated pushes}
  \label{tab:rep_push}
  \centering
  \begin{tabular}{|l|r|r|r|} \hline
    \bfseries Surface  & \bfseries Mean & \bfseries Trans. std & \bfseries Rot. std \\
    & \bfseries (mm, mm, deg) & \bfseries (mm)    & \bfseries (deg)    \\ \hline
    \texttt{abs}       & (40.1, -67.6, 74.7) & 5.5 (7.1\%) & 3.2 (4.3\%) \\ \hline
    \texttt{delrin}    & (38.8, -50.7, 78.5) & 3.4 (5.2\%) & 1.3 (1.6\%) \\ \hline
    \texttt{plywood}   & (36.4, -93.6, 70.2) & 8.1 (8.0\%) & 4.2 (6.0\%) \\ \hline
    \texttt{pu}        & (40.2, -85.0, 69.3) & 11.7 (12.5\%) & 4.5 (6.5\%) \\ \hline\hline
    simulator \cite{lynch1992manipulation}   & (41.0, -98.1, 66.3) & N/A & N/A \\ \hline
  \end{tabular}
\end{table}

\figref{fig:rep_push_viz_abs} also shows a comparison of the mean experimental trajectory and the prediction by a model driven simulator \cite{lynch1992manipulation}. They look quite different. Thus, another interesting future direction is to better evaluate those differences, in particular, under what conditions the predictions of a simple deterministic model are reasonable. 

%
Experiments show that even when trying to replicate the same initial conditions with an accurate vision system and an accurate robot, a determined pushing interaction yields appreciable and structured uncertainty at the outcome. This motivates further investigation of effective ways to take into account uncertainty or variability in friction.


%
%
%

\begin{figure}
  \begin{center}
    \includegraphics[width=3.4in]{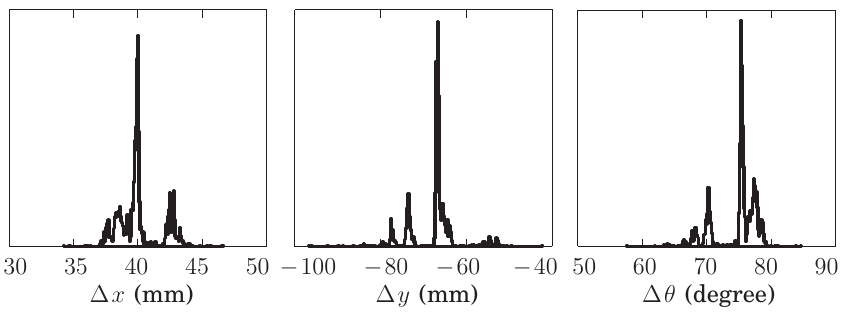}
  \end{center}
  \caption{Histogram of displacements $\Delta x$, $\Delta y$, and $\Delta \theta$ produced by the 2000 pushes in \figref{fig:rep_push_viz_abs}.}
  \label{fig:rep_push_histogram}
\end{figure}

\section{Conclusion}
\label{sec:conclusion}

This paper presents a large and high-fidelity experimental dataset of planar pushing interactions. The data spans six different dimensions of the pushing problem: the shape of the pushed object, the material of the surface where it slides, the location of contact between pusher and slider, and the direction, velocity, and acceleration along which the pusher moves. Overall, these generate more than a million timestamped samples of positions of pusher and slider, as well as interaction forces.

We also describe and evaluate the most common assumptions and approximations used in models of planar pushing. The results say that while some assumptions such as the maximum power inequality are generally good representations of the relationship between the directions of friction and motion, other assumptions are not equally respected. In particular the ratio between the magnitudes of normal force and friction force at contact (i.e. the coefficient of friction) is not necessarily constant,
and the ratio changes in space, with orientation, with velocity, and with time. As expected, materials that are harder yield slightly better approximations. 

Our current and future work include leveraging this dataset to develop more accurate semi-parametric and stochastic models of frictional pushing;  investigating its use in the context of simulation, planning, and control; as well as continued efforts in collecting experimental data for prehensile~\citep{Kolbert2016} and non-prehensile~\citep{fazeli2015drop} contact interactions. Of particular interest are out-of-plane motions that involve different manipulation actions such as rolling or toppling.

Our long term goal is to steer away from a manipulation paradigm that relies heavily on open loop executions of motions that are planned with simple deterministic models of frictional interaction.

\bibliographystyle{IEEEtranN} 
{\footnotesize \bibliography{iros16-pushing}} 

\end{document}